\documentclass[journal]{IEEEtran}
\usepackage{bm}
\usepackage{amsmath}
\usepackage{amssymb}
\usepackage{amsthm}
\usepackage{mathrsfs}
\usepackage{enumerate}
\usepackage{multirow}
\usepackage{color}
\usepackage{subfigure}
\usepackage[square, comma, sort&compress, numbers]{natbib}
\usepackage[pagebackref=false,breaklinks=true,letterpaper=true,colorlinks,bookmarks=false]{hyperref}
\usepackage{times}
\usepackage{breakurl}
\usepackage{array}
\usepackage{verbatim}
\usepackage{algorithm}
\usepackage{algpseudocode}
\definecolor{orange}{rgb}{1,0.5,0} 
\definecolor{purple}{rgb}{0.5,0,0.5} 

\makeatletter
\def\hlinewd#1{%
  \noalign{\ifnum0=`}\fi\hrule \@height #1 \futurelet
   \reserved@a\@xhline}
\makeatother

\ifCLASSINFOpdf
  \usepackage[pdftex]{graphicx}
  \usepackage{graphics}
  \usepackage{graphicx}
  \usepackage{epsfig}
  \usepackage{epstopdf}
  \graphicspath{{./fig/}}
  \DeclareGraphicsExtensions{.pdf,.jpeg,.png,.eps}
\else
  \usepackage[dvips]{graphicx}
  \graphicspath{{./fig/}}
  \DeclareGraphicsExtensions{.eps}
\fi

\begin{document}

\title{Exploring Spatial Significance via Hybrid Pyramidal Graph Network for Vehicle Re-identification}
\author{Fei Shen, Jianqing Zhu, Xiaobin Zhu, Yi Xie, and Jingchang Huang
\thanks{
	
This work was supported in part by the National Natural Science Foundation of China under the Grant 61976098, in part by the Natural Science Foundation of Fujian Province under the Grant 2018J01090, in part by the Open Foundation of Key Laboratory of Security Prevention Technology and Risk Assessment, People's Public Security University of China under the Grant 18AFKF11, in part by the Science and Technology Bureau of Quanzhou under the Grant 2018C115R, in part by the Promotion Program for Young and Middle-aged Teacher in Science and Technology Research of Huaqiao University under the Grant ZQN-PY418, and in part by the Scientific Research Funds of Huaqiao University under the Grant 16BS108 \emph{(Corresponding author: Jianqing Zhu and Jingchang Huang).
}}
\thanks{Fei Shen, Jianqing Zhu, and Yi Xie are with Huaqiao University, Quanzhou, 362021, China (e-mail: feishen@stu.hqu.edu.cn, jqzhu@hqu.edu.cn, and yixie@stu.hqu.edu.cn).}
\thanks{Xiaobin Zhu is School of Computer and Communication Engineering, University of Science and Technology Beijing,Xueyuan Road 30, Haidian District, Beijing, 100083 China (e-mail: zhuxiaobin@ustb.edu.cn).}
\thanks{Jingchang Huang is with Shanghai Institute of Micro-system and Information Technology, Chinese Academy of Sciences, No. 865 Changning Road, Shanghai, 200050, China (e-mail:jchhuang@mail.ustc.edu.cn).}
}

\markboth{}%
{Shell \MakeLowercase{\textit{et al.}}: Bare Demo of IEEEtran.cls for Journals}

\maketitle

\begin{abstract}
Existing vehicle re-identification methods commonly use spatial pooling operations to aggregate feature maps extracted via off-the-shelf backbone networks. They ignore exploring the spatial significance of feature maps, eventually degrading the vehicle re-identification performance. In this paper, firstly, an innovative spatial graph network (SGN) is proposed to elaborately explore the spatial significance of feature maps. The SGN stacks multiple spatial graphs (SGs). Each SG assigns feature map's elements as nodes and utilizes spatial neighborhood relationships to determine edges among nodes. During the SGN's propagation, each node and its spatial neighbors on an SG are aggregated to the next SG. On the next SG, each aggregated node is re-weighted with a learnable parameter to find the significance at the corresponding location. Secondly, a novel pyramidal graph network (PGN) is designed to comprehensively explore the spatial significance of feature maps at multiple scales. The PGN organizes multiple SGNs in a pyramidal manner and makes each SGN handles feature maps of a specific scale. Finally, a hybrid pyramidal graph network (HPGN) is developed by embedding the PGN behind a ResNet-50 based backbone network. Extensive experiments on three large scale vehicle databases (i.e., VeRi776, VehicleID, and VeRi-Wild) demonstrate that the proposed HPGN is superior to state-of-the-art vehicle re-identification approaches.

\end{abstract}


\IEEEpeerreviewmaketitle

\section{Introduction}
	Vehicle re-identification \cite{veri2016,veri2017} is a challenging yet meaningful computer vision task of retrieving vehicle images holding the same identity but captured from different surveillance cameras. Vehicles have different appearances at different spatial locations, which naturally causes feature maps containing varying spatial significance. Especially, subtle cues in local regions (e.g., lightings, stickers or pendants on windshields, and license plates) play critical roles in identifying highly similar vehicles of the same color and type/model. Therefore, exploring spatial significance of feature maps is crucial to improve vehicle re-identification performance.

    In general, deep learning based vehicle re-identification models consist of backbone networks and feature aggregation architectures. Backbone networks take charge of learning three-dimensional (i.e., $height\times width\times channel$) feature maps from vehicle images. Feature aggregation architectures are responsible for aggregating the learned three-dimensional feature maps into feature vectors. Regarding backbone networks, a lot of off-the-shelf deep networks, such as VGGNet \cite{vgg}, GoogLeNet \cite{googlenet}, ResNet \cite{resnet,resnetv2}, DenseNet \cite{dense}, and MobileNet \cite{mobnet}, can be adopted. Based on those networks, backbone networks hold strong feature learning abilities. However, they still lack a special mechanism to explore the spatial significance of feature maps explicitly. One can design a new backbone network without using off-the-shelf deep networks to explore the spatial significance of feature maps. Nevertheless, it is challenging and daunting to design a new backbone network that can outperform off-the-shelf networks. Especially, those off-the-shelf deep networks have shown excellent advantages in many computer vision tasks. Therefore, how to explore the spatial significance of feature maps by feature aggregation architectures is received more and more research interests recently.

	For aggregating feature maps extracted via backbone networks, many vehicle re-identification methods (e.g., \cite{veri2017, mgr, aaver, imtri, drdl,djdl, mattribute, vanet,vst,c2f,triemd,vric,msv,ealn,dban}) use spatial global pooling operations. Although spatial global pooling operations are beneficial to learning viewpoint robust features, they only calculate feature maps' spatial statistics (i.e., maximum or average), but ignore to explore the spatial significance of feature maps. To explore the spatial significance of feature maps, an alternative way is to divide vehicles into several parts and then features are learned on each part individually. Regarding part division ways, there are (1) uniform spatial divisions \cite{prn,qddlf,jquad,ram,san,sff}, (2) spatial divisions \cite{rpm,tamr} using the spatial transformer network \cite{stn}, (3) part detector based spatial divisions \cite{partreg,pgan,pmsm}, and (4) key-point regressor based divisions \cite{oife,aaver,attdriven,pamtri}. The first two kinds of spatial division methods are computationally economical but prone to suffer from part misalignments. In contrast, the last two kinds of spatial division methods could solve dis-alignments but encounter a high cost of extra manual part/key-point annotations and expensive training/inference computations. However, no matter how to divide parts, the subsequent feature learning is individually implemented on each part. As a result, the spatial significance of feature maps could not be jointly explored across multiple parts.
	
\begin{figure}[tp]
	\centering
	\includegraphics[width=1\linewidth]{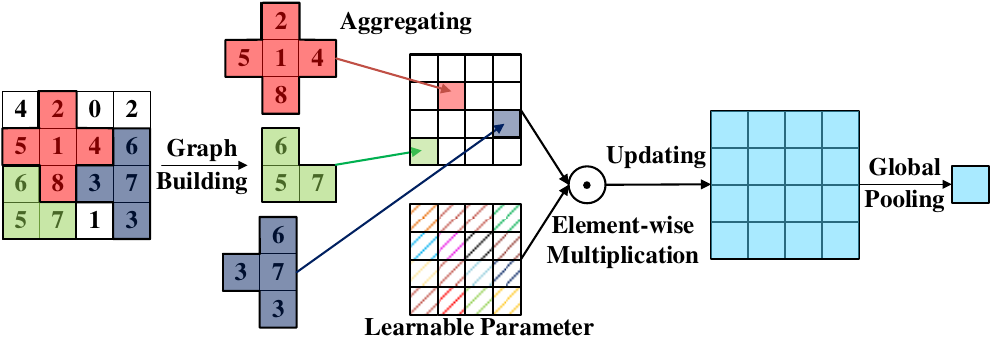}
	\caption{ {\color{black}
			Exploring  spatial significance before global pooling.
			For an easy visualization, single-channel feature maps are presented.
		}
	}
	\label{fig:example}
\end{figure}

	In this paper, to elaborately explore the spatial significance of feature maps extracted via a backbone network, an innovative spatial graph network (SGN) is designed. The SGN builds spatial graphs on the feature maps and re-weights feature maps' elements at each spatial location via a learnable parameter before average pooling, as shown in Fig. \ref{fig:example}.
	Furthermore, to comprehensively explore the spatial significance of feature maps at multiple scales, a novel pyramidal graph network (PGN) is designed to organize multiple SGNs and make each SGN deals with feature maps of a specific scale. As a result, a hybrid pyramidal graph network (HPGN) exploring spatial significance for vehicle re-identification is proposed by embedding the proposed PGN behind a ResNet-50 \cite{resnet} based backbone network.

	The main contributions of this paper are summarized in three folds. (1) An innovative spatial graph network (SGN) is designed to elaborately explore the spatial significance of feature maps resulted from a backbone network, without requiring any extra
	part division operations or part detectors/key-point regressors. To the best of our acknowledge, it is the first attempt to explore the spatial significance via graph networks. (2) A novel pyramidal graph network (PGN) is proposed to organize multiple SGNs to comprehensively explore the spatial significance of feature maps at multiple scales. (3) Extensive experiments on three large scale vehicle databases (i.e., VeRi776 \cite{veri776}, VehicleID \cite{drdl}, and VeRi-Wild \cite{wild}) show that the proposed method is superior to many state-of-the-art vehicle re-identification approaches.
	
	The rest of this paper is organized as follows. Section \ref{sec:rw} introduces the related work. Section \ref{sec:method} describes the proposed method. Section \ref{sec:exp} presents experiments and analyses to validate the proposed method's superiority. Section \ref{sec:con} concludes this paper.

\section{Related Work}\label{sec:rw}
\subsection{Vehicle Re-identifications}
	Given a backbone network based on off-the-shelf backbone networks (e.g., VGGNet \cite{vgg}, GoogLeNet \cite{googlenet}, ResNet \cite{resnet,resnetv2}, DenseNet \cite{dense}, and MobileNet \cite{mobnet}), vehicle re-identification progress is reviewed from two aspects: (1) feature aggregation architectures and (2) loss functions.
	
	\subsubsection{Feature Aggregation Architectures}
	There are two types of feature aggregation architectures: (1) the global-level feature aggregation architecture and (2) the part-level feature aggregation architecture. The global-level aggregation architecture usually is a spatial global pooling layer, as done in \cite{veri2017, mgr, aaver, imtri, drdl,djdl, mattribute, vanet,vst,c2f,triemd,vric,msv,ealn,dban}. Spatial global pooling layers are non-parameterized, which find the maximal or average spatial statistic on each channel of feature maps extracted via a backbone network. Using spatial global pooling layers, there is an advantage of learning viewpoint robust features. However, due to the characteristic of spatial global pooling layers, the varying spatial significance of feature maps will be inevitably ignored, which is detrimental to vehicle re-identification.
	
    According to the way of dividing vehicle parts, there are two kinds of part-level feature aggregation architectures: (1) part annotation free feature aggregation architectures and (2) part annotation required feature aggregation architectures. Regarding part annotation free feature aggregation architectures, the most straightforward and economical method is the uniform spatial division pooling strategy, which equally divides feature maps into several parts and pools each part individually, as done in \cite{prn,qddlf,jquad}. Since parts are divided rigidly, the uniform spatial division pooling strategy is prone to suffering from part dis-alignments. To alleviate dis-alignments, \cite{ram,san,sff} simultaneously apply spatial global and spatial division pooling strategies, and \cite{rpm,tamr} use the spatial transformer network \cite{stn} to determine part regions. Besides, in \cite{sff, rpm,tamr}, visual attention modules(e.g., the self-attention \cite{selfatt} and the residual attention \cite{spatialattention}) are adopted for refining features on each part. Nevertheless, part dis-alignments are still serious since lacking accurate part detectors.
	
	Regarding part annotation required feature aggregation architectures, in \cite{partreg,pgan,pmsm}, typical detectors, such as the faster region-based convolutional neural network (Faster R-CNN) \cite{fasterrcnn}, the you only look once (YOLO) model \cite{yolo}, and the single shot multi-box detector (SSD) \cite{ssd}, are applied to detect vehicle parts, e.g., lightings, front windshields, and logos. Additionally,  in \cite{oife, aaver,attdriven,pamtri}, key-point regressors are employed to locate informative vehicle regions. However, both typical detectors and key-point regressors are deep networks in themselves, along with a high cost of extra manual part annotations or key-point annotations and expensive training/inference computations.

\begin{figure*}[htp]
	\centering
	\includegraphics[width=.95\linewidth]{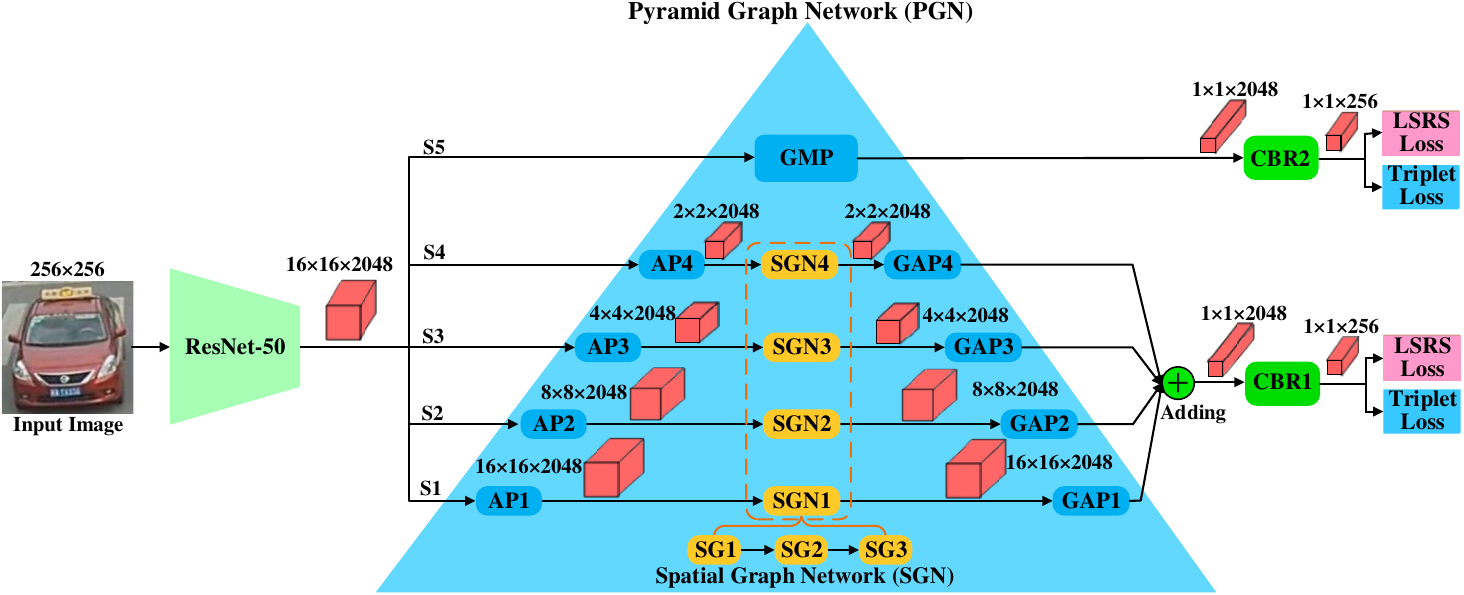}
	\caption{
		{\color{black}
		The framework of the proposed hybrid pyramidal graph network (HPGN). Here,  AP denotes an average pooling layer; GAP and GMP respectively represent a global average pooling layer and a global max pooling layer; CBR represents a composite unit that consists of a convolutional layer using $1\times1$ sized filters, a batch normalization layer \cite{bnorm} and a rectified linear unit (ReLU) \cite{relu}; LSRS means that the label smoothing regularized softmax loss \cite{LSR}.
	}
	}
	\label{fig:framework}
\end{figure*}

\subsubsection{Loss Functions}
Many vehicle re-identification methods (e.g., \cite{triemd,imtri,drdl,djdl,tamr}) show promising performance by combining softmax \cite{deepid,LSR} and triplet loss \cite{facenet, imtri} functions. Because the softmax loss is beneficial to the convergence speed, and the triplet loss is helpful to learn discriminative features. In addition to contaminating softmax \cite{deepid} and triplet \cite{imtri} loss functions, sample mining strategies (e.g., hard negative/positive sample mining) also can improve vehicle re-identification \cite{triemd}. Besides, there are some variant triplet loss functions \cite{imtri,drdl,ggl,gste, mrl,vanet} for vehicle re-identification. For example, the group-group (GGL) loss function \cite{ggl} computes the loss on image groups rather than image pairs or triplets, simplifying the model training process. The group sensitive triplet embedding (GSTE) method \cite{gste} improves the triplet loss function by well dealing with the intra-class variance. The multi-view ranking loss (MRL) \cite{mrl} approach divides the triplet loss into a single-view ranking loss and a cross-view ranking loss to appropriately handle viewpoint variations.

\subsection{Graph Convolutional Network}
The graph convolutional network (GCN) \cite{gcn, gcnp, gcnc} initially is for learning features on non-Euclidean data, which can flexibly aggregate nodes to learn features. Recent studies show that GCN has high potentials in many computer vision tasks, such as multi-label recognition \cite{mlir}, action recognition \cite{stgcn, twogcn, asgcn}, and person re-identification \cite{shengcn, dotgnn, mgan, dynamicgcn, structuregcn}. Especially, with the strength of GCN, a similar task to this paper, i.e., person re-identification, receives a list of advantages, such as the refining of the similarity between an image pair \cite{shengcn}, the robustness to individuals’ appearance and membership changes \cite{dotgnn}, and the robustness to noise of initial training data \cite{dynamicgcn}.

Regarding the graph construction way, all those GCN based person re-identification methods \cite{shengcn, dotgnn, mgan, dynamicgcn, structuregcn} build a graph involving multiple individuals where an individual person is assigned as a node. In contrast, this paper independently constructs a graph for each individual. More specifically, this paper develops a graph network on feature maps of a vehicle individual to elaborately explore the spatial significance. In the proposed graph network, each element of a feature map is assigned as a node, and a parameter is learned for re-weighting each node to explore the spatial significance of feature maps.

\section{Proposed Method}\label{sec:method}
\subsection{Overview}
	As shown in Fig. \ref{fig:framework}, the proposed hybrid pyramidal graph network (HPGN) is mainly constructed by embedding a pyramidal graph network (PGN) following a ResNet-50 based backbone network. Regarding the PGN, it is composed of multiple spatial graph networks (SGNs). Each SGN stacks three spatial graphs (SGs) to handle a feature map of a specific scale. Each SG assigns feature map's elements as nodes and utilizes spatial neighborhood relationships to determine edges among nodes. Each node and its spatial neighbors on an SG are aggregated to the next SG during the SGN's propagation. Then, each aggregated node of the next SG is re-weighted with a learnable parameter to automatically find the significance at the corresponding location. Regarding the backbone network, the ResNet-50 \cite{resnet} is applied, which is commonly used in previous re-identification works \cite{luo2019strong,mgl,bsvr,partreg,aaver}. Besides, following \cite{luo2019strong}, the `last stride=1' training trick is applied to set the last stride of the fourth residual layer to 1. Consequently, for a $256\times256$ sized input vehicle image, the ResNet-50 extracts a $16\times16\times2048$ sized feature map that will be further handled with the PGN and loss functions. More detail about the PGN and loss functions are described as follows.

\subsection{Pyramidal Graph Network}
	The PGN contains two essential designs, namely, the spatial graph network (SGN) and the pyramidal architecture (PA). The SGN is designed to explore the spatial significance of feature maps at one scale. The PA is designed to organize multiple SGNs to comprehensively explore spatial significance of feature maps at multiple scales. The detail of SGN and PA are introduced as follows.
	
\subsubsection{Spatial Graph Network}
	Assume that $X \in R^{h\times w\times c}$ is a three-dimensional  feature map extracted by using the backbone network (i.e., ResNet-50 \cite{resnet}), where $h$, $w$, and $c$ represents height, width, and channel number of $X$. For ease of description, $X$ is reshaped as a matrix $V\in R^{d\times c}$ by rearranging each channel of $X$ into a $d$ dimensional vector, where $d=h\times w$. Then, a spatial graph $G = \{V \in R^{d\times c} , E \in R^{d\times d \times c}\}$ assigns each element of the vector $V$ as a node. Furthermore, the spatial graph's edges are constructed according to Eq. \eqref{eq:edge}, as follows.
	\begin{equation}\label{eq:edge}
	E_{i,j,k}=\left\{
	\begin{array}{cc}
	\frac{1}{|N_{V_{i,k}}|}, & V_{j,k} \in N_{V_{i,k}}, \\
	0, & {otherwise},
	\end{array}\right.
	\end{equation}
	where $i, j \in [1,2,...,d]$ and $k \in [1,2,...,c]$; $E_{i,j,k}$ is the edge between the node $V_{i,k}$ and the node $V_{j,k}$ on the $k$-th channel; $N_{V_{i,k}}$ represents the neighbor node set of $V_{i,k}$, and $|N_{V_{i,k}}|$ denotes the number of nodes in $N_{V_{i,k}}$.
	
	In order to simplify the edge definition of Eq. \eqref{eq:edge}, the edge definition on each channel is assumed to be identical. Specifically, $N_{V_{i,k}}$ is determined according to the four spatial neighbors (i.e., up, down, left, and right) of node $V_{i,k}$. Moreover, if $V_{i,k}$ is a boundary node, its neighbors are reduced according to its actual location since zero-padding operations are given up in this paper. As a result, the edge definition of Eq. \eqref{eq:edge} is simplified as follows:
	\begin{equation}\label{eq:edge2}
	\begin{array}{cc}
	{E_{i,j,1}} = {E_{i,j,2}} = ... = {E_{i,j,c}} = {S_{i,j}},\\
	{S_{i,j}} = \left\{ {\begin{array}{*{20}{c}}
		{\frac{1}{{|{N_{{V_{i,1}}}}|}},}&{{V_{j,1}} \in {N_{{V_{i,1}}}},}\\
		{0,}&{otherwise.}
		\end{array}} \right.\\
	\end{array}
	\end{equation}
	
	Based on the simplified edge definition of Eq. \eqref{eq:edge2}, the spatial graph $G = \{V \in R^{d\times c} , S \in R^{d\times d}\}$ is built, and the node aggregation is according to Eq. \eqref{eq:agg}, as follows.
	\begin{equation}\label{eq:agg}
	A=IV+SV=\left(I+S\right) V,
	\end{equation}
	where $I=diag(1,1,1,...,1) \in R^{d\times d}$ is an identity matrix; $A\in R^{d\times c}$ is a matrix containing aggregated nodes.
	Then, each aggregated node is further updated by using Eq. \eqref{eq:dot}, as follows:
	\begin{equation}\label{eq:dot}
	U=A\odot {\varTheta}, 
	\end{equation}
	where $\odot$ represents an element-wise multiplication operation; $\varTheta \in R^{d\times c}$ is a learnable parameter matrix used to explore the spatial significance via re-weighting aggregated nodes of $A$. Consequently, $U$ can take the varying spatial significance of feature maps into account.
	
	From Eq. \eqref{eq:agg} and Eq. \eqref{eq:dot}, one can see that both a node itself and its neighbor nodes are
	aggregated to feed more information for exploring the significance at the corresponding location. Besides, regarding Eq. \eqref{eq:dot}, different from existing works \cite{shengcn,dotgnn} that use a multi-layer perceptron (MLP) to update aggregated nodes, the proposed method applies an element-wise multiplication operation. Because the goal of proposed method is to explore the spatial significance rather than to transform features complexly.

	Both Eq. \eqref{eq:agg} and Eq. \eqref{eq:dot} are linear calculations. In order to introduce non-linearities and improve the spatial graph's convergence, Eq. \eqref{eq:dot} is
	enhanced as follows:
	\begin{equation}\label{eq:nonlinear}
	O=LeakyReLU \left(BN\left(U\right)\right) ,
	\end{equation}
	where $LeakyReLU$ represents the leaky rectified linear unit \cite{leakrelu} of a 0.2 slope parameter; $BN$ denotes the batch normalization \cite{bnorm}; $O \in R^{d\times c}$ replaces $U$ as the new feature map that non-linearly deals with the spatial significance. Finally, according to Eq. \eqref{eq:edge2}, Eq. \eqref{eq:agg}, Eq. \eqref{eq:dot}, and Eq. \eqref{eq:nonlinear}, a spatial graph (SG) can be built, and then the spatial graph network (SGN) is constructed by stacking multiple spatial graphs (SGs).

\subsubsection{Pyramidal Architecture}
	As shown in Fig. \ref{fig:framework}, a pyramidal architecture containing multiple SGNs is further designed to comprehensively explore the spatial significance of feature maps at multiple scales. First, for a $16\times 16 \times 2048$ sized feature map extracted by the ResNet-50 \cite{resnet} based backbone network, five pooling layers (i.e., AP1, AP2, AP3, AP4, and GMP) are simultaneously applied to resize the feature map to acquire multi-scale feature maps, i.e., $16\times 16 \times 2048$, $8\times 8\times 2048$, $4\times 4\times 2048$, $2\times 2\times 2048$, and $1\times 1\times 2048$ sized feature maps. The configurations of these five pooling layers are listed in Table \ref{tab:config}. Please note that the AP1 layer is just for the elegant presentation of Fig. \ref{fig:framework}. In practice, the AP1 layer can be removed since performing an $1\times1$ sized average pooling operation on a feature map does not change the feature map.
	
	Second, following AP1, AP2, AP3, and AP4 layers, four spatial graph networks (i.e., SGN1, SGN2, SGN3, and SGN4) are applied to explore the spatial significance of $16\times 16 \times 2048$, $8\times 8\times 2048$, $4\times 4\times 2048$, $2\times 2\times 2048$ sized feature maps, respectively. Moreover, on the minimal sized (i.e., $1\times1\times2048$) feature map,  there is no spatial graph network since the $1\times1$ resolution is too small.
	
	Third, following SGN1, SGN2, SGN3, and SGN4, four global average pooling (i.e., GAP1, GAP2, GAP3, and GAP4) are respectively employed to aggregate feature maps. The configurations of  GAP1, GAP2, GAP3, and GAP4 are also listed in Table \ref{tab:config}. In our implementation, to reduce model complexities, these four spatial graph networks (SGNs) hold the same architecture, i.e., each SGN has three spatial graphs, as shown in Fig. \ref{fig:framework}.

	\begin{table}[tp]
		\caption{The configuration of pooling layers in the proposed PGN.
		} \label{tab:config}
		\small
		\setlength{\tabcolsep}{3pt}
		\begin{center}
			\begin{tabular}{ccccc}
				\hline
				Components  &
				\begin{tabular}{c}
					Pooling Window\\
					($height \times width$)\\
				\end{tabular}
				& Channels
				& Stride
				& Padding \\ \hline
				AP1               & $1 \times 1$         & 2048                                                    & 1 & 0\\
				AP2               & $2 \times 2$         & 2048                                                    & {2} & 0    \\
				AP3               & $4 \times 4$         & 2048                                                    & {4} & 0    \\
				AP4               & $8 \times 8$         & 2048                                                    & {8} & 0    \\
				GMP             & $16\times16$      & 2048                                                    & 1  & 0\\
				GAP1            & $16 \times 16$    & 2048                                                    & 1  & 0    \\
				GAP2            & $8 \times 8$         & 2048                                                    & 1  & 0   \\
				GAP3            & $4 \times 4$         & 2048                                                    & 1  & 0   \\
				GAP4            & $2 \times 2$         & 2048                                                    & 1  & 0\\
				\hline
			\end{tabular}
		\end{center}
		\vspace{-.4cm}
	\end{table}

\subsection{Loss Function Design}
	From Fig. \ref{fig:framework}, one can see that the proposed PGN applies a global max pooling (GMP) layer to produce the minimal sized (i.e., $1\times1\times2048$) feature map, while uses average pooling layers (i.e., GAP1, GAP2, GAP3, and GAP4) to generate larger sized feature maps. This configuration of using two kinds of pooling layers are beneficial to enrich multi-scale feature maps. Based on this configuration, two sets of loss functions are assigned to supervise features resulted from max and average pooling layers. More details are described as follows.
	
	Firstly, an adding layer is utilized to accumulate features produced by GAP1, GAP2, GAP3, and GAP4. Secondly, two CBR composite units (i.e., CBR1 and CBR2) are embedded after the adding layer and the GMP layer individually to reduce the feature dimension from 2048 to 256. The CBR composite unit is composed of a convolutional layer using $1\times1$ sized filters, a batch normalization \cite{bnorm} layer, and a rectified linear unit (ReLU) \cite{relu}. Finally, on CBR1 and CBR2, the proposed HPGN's total loss function is formulated as follows:
	\begin{equation}\label{eq:totoal}
	{L_{total}} = \alpha L_{lsrs}^{CBR1} + \beta L_{triplet}^{CBR1}+
	\rho L_{lsrs}^{CBR2}+ \lambda L_{triplet}^{CBR2},
	\end{equation}
	where $L_{lsrs}^{CBR1}$ and $L_{lsrs}^{CBR2}$ are label smoothing regularized softmax (LSRS) loss functions \cite{LSR} playing on CBR1 and CBR2, respectively; similarly, $L_{triplet}^{CBR1} $ and $L_{triplet}^{CBR2}$ are the triplet loss functions \cite{triplet} working on CBR1 and CBR2, respectively;  $\alpha$, $\beta$, $\rho$, and $\lambda \ge 0$ are manually setting constants used to keep the balance of four loss functions.
	To avoid excessive tuning those constants, we set $\alpha=\rho=2$ and $\beta=\lambda = 1$ in the following experiments. The LSRS loss function \cite{LSR} is formulated as follows:
	\begin{equation}\label{eq:llsrs}
	{
		{L_{lsrs}}(X,l) = \frac{{ - 1}}{M}\sum\limits_{i = 1}^M {\sum\limits_{j = 1}^K {\delta ({l_i},j)log(\frac{{{e^{W_j^{\rm{T}}{X_i}}}}}{{\sum\nolimits_{k = 1}^K {{e^{W_k^{\rm{T}}{X_i}}}} }})} },
	}
	\end{equation}
	\begin{equation}\label{eq:delta}
	{
		\delta ({l_i},j) =\left\{
		\begin{array}{cc}
		1 - \varepsilon  + \frac{\varepsilon }{K}, & j = {l_i}, \\
		\frac{\varepsilon }{K}, & {otherwise},
		\end{array}\right.
	}
	\end{equation}
	where $X$ is a training set and $l$ is the class label information; $M$ is the number of training samples; $K$ is the number of classes; $(X_i, l_i)$ is the $i$-th training sample and $l_i \in \{1, 2, 3, ..., K\}$; $W=[W_1, W_2, W_3, ..., W_K]$ is a learnable parameter matrix; $\varepsilon \in [0,1)$ is a tiny constant used to control the smoothing regularization degree, which is set to $0.1$ in this paper, as done in \cite{LSR}.
	
	The triplet loss function \cite{triplet} is formulated as follows:
	\begin{equation}\label{eq:triplet}
	{
		\begin{aligned}
		&{L_{triplet}}({X^a},{X^n},{X^p}) \\=
		&\frac{{ - 1}}{M}\sum\limits_{i = 1}^M {\max ({{\left\|{X_i^a - X_i^n} \right\|}_2}-{{\left\|  {X_i^a - X_i^p} \right\|}_2} - \tau,0)},
		\end{aligned}
	}
	\end{equation}
	where $({X^a},{X^n},{X^p})$ is a set of training triplets; $M$ is the number of training triplets; for the $i$-th training triplet, $(X^a_i, X^n_i)$ is a negative pair of different class labels, and $(X^a_i, X^p_i)$ is a positive pair of the same class label; $\tau\geq0$ is a margin constant, which is set to $1.2$ in following experiments; $\left\|\cdot\right\|_2$ denotes the Euclidean distance. Besides, the hard sample exploring strategy \cite{triplet} is applied to improve the triplet loss, aiming to find the most difficult positive and negative image pairs in each mini-batch.

{
\section{Experiment and Analysis}\label{sec:exp}
    To validate the superiority of the proposed hybrid pyramidal graph network (HPGN) method, it is compared with state-of-the-art vehicle re-identification approaches on three large-scale databases.} The rank-1 identification rate (R1)  and {mean average precision} (mAP)  are used as performance metrics.

\subsection{Databases}
     \textbf{{VeRi776}}~\cite{veri776} is collected by 20 cameras in unconstrained traffic scenarios, and each vehicle is captured by 2-18 cameras. Following the evaluation protocol of \cite{veri776}, VeRi776 is separated into a training subset and a testing subset. The training subset contains 37,746 images of 576 subjects. The testing subset includes a probe subset of 1,678 images of 200 subjects and a gallery subset of 11,579 images of the same 200 subjects. Besides, only cross-camera vehicle pairs are evaluated, which means that if a probe image and a gallery image are captured by the same camera, the corresponding result will be excluded in the evaluation process.

     \textbf{{VehicleID}} \cite{drdl} includes 221,763 images of 26,267 subjects. Each vehicle is captured from either front or rear viewpoint. The training subset consists of 110,178 images of 13,164 subjects. There are three testing subsets, i.e., Test800, Test1600, and Test2400, for evaluating the performance at different data scales. Specifically, Test800 includes 800 gallery images and 6,532 probe images of 800 subjects. Test1600 contains 1,600 gallery images and 11,395 probe images of 1,600 subjects. Test2400 consists of 2,400 gallery images and 17,638 probe images of 2,400 subjects. Besides, for three testing subsets, the division of probe and gallery subsets is as follows: randomly selecting one image of a subject to form the probe subset, and the subject's remaining images are used to construct the gallery subset. This division is repeated and evaluated ten times, and the average result is reported as the final performance.

     \textbf{{VERI-Wild}}~\cite{wild} is a newly dataset released in CVPR 2019. Different to VeRi776 \cite{veri776} and VehicleID \cite{drdl} captured at day, VERI-Wild are captured at both day and night. The training subset of 277,797 images of 30,671. Besides, as in VehicleID \cite{drdl}, there are three different scale testing subsets, i.e., Test3000, Test5000, and Test10000. Test3000 has 41,816 gallery images and 3000 probe images of 3,000 subjects. Test5000 contains 69,389 gallery images and 5,000 probe images of 5,000 subjects. Test10000 includes 138,517 gallery images and 10,000 probe images of 10,000 subjects.

\vspace{-.2cm}
\subsection{Implementation Details}
    The deep learning toolbox is Pytorch \cite{pytorch}. Training configurations are summarized as follows. (1) Random erasing \cite{RS}, horizontal flip, and z-score normalization are used for the data augmentation. For both horizontal flip and random erasing operations, the implementation probability is set to 0.5. (2) The mini-batch stochastic gradient descent (SGD) method \cite{alexnet} is applied to optimize parameters. The weight decays are set to 5$\times$10$^{-4}$, and the momentums are set to 0.9. There are 150 epochs for the training process. The learning rates are initialized to 3$\times$10$^{-4}$, and they are linearly warmed up \cite{luo2019strong} to 3$\times$10$^{-2}$ in the first 10 epochs. After warming up, the learning rates are maintained at 3$\times$10$^{-2}$ from 11th to 50th epochs. Then, the learning rates are reduced to 3$\times$10$^{-3}$ from 51st to 85th epochs. Furthermore, the learning rates are decayed to 3$\times$10$^{-4}$ after 85 epochs, and they are reduced to 3$\times$10$^{-5}$ from 120th and 150th epochs. (3) Each mini-batch includes 32 subjects, and each subject holds 4 images. During the testing phase, those 256-dimensional features resulted from CBR1 and CBR2 (see Fig. \ref{fig:framework}) are concatenated as final features of vehicle images. Moreover, the Cosine distance of final features is applied as the similarity measurement for vehicle re-identification.


\subsection{Comparison with State-of-the-art Methods}
\begin{table}[tp]
	\caption{The performance (\%) comparison on VeRi776. The {\color{red}{red}}, {\color{green}{green}} and {\color{blue}{blue}} rows respectively represent the {\color{red}{$1$st}}, {\color{green}{$2$nd}} and {\color{blue}{$3$rd}} places, according to the R1 comparison.
	} \label{tab:veri776}
	\begin{center}
		\footnotesize
		\setlength{\tabcolsep}{2pt}
		\begin{tabular}{lccr}
			\hline
			{Methods} &R1  & mAP  & References \\
			\hline
			{\color{red}{HPGN}}
			&{\color{red}{96.72}}
			&{\color{red}{80.18}}
			& {\color{red}{Proposed}}\\ 
			QD-DLF \cite{qddlf}                                                     &88.50&61.83&  IEEE ITS 2020\\
			\hline
			
			{\color{green}{Appearance+License \cite{app}}  } &{\color{green}{95.41}}
			&{\color{green}{78.08}}
			&{\color{green}{ICIP 2019}} \\
			
			{\color{blue}{SFF+SAtt \cite{sff}}}
			& {\color{blue}{94.93}}
			& {\color{blue}{74.11}}
			& {\color{blue}{IJCNN 2019}}\\
			Part Regularization \cite{partreg}
			&94.30&74.30   & CVPR 2019\\
			SAN \cite{san} &93.30 &72.50   & arXiv 2019\\
			PAMTRI \cite{pamtri} & 92.86 & 71.88   & ICCV2019\\
			MLFN+Triplet \cite{mlfn} &92.55 & 71.78   & CVPRW 2019\\
			MTML+OSG+Re-ranking \cite{mtml} &92.00 &68.30   & CVPRW 2019\\
			MRM \cite{mrm} &91.77 &68.55   & Neurocomputing 2019\\
			DMML \cite{dmml} &91.20 &70.10   & ICCV 2019\\
			Triplet Embedding \cite{triemd}                  &90.23&67.55  & IJCNN 2019\\
			MOV1+BS \cite{cityflow} & 90.20 & 67.60   & CVPR 2019\\
			VANet \cite{vanet}                  &89.78&66.34   & ICCV 2019\\
			GRF+GGL \cite{ggl} &89.40 &61.70   & IEEE TIP 2019\\
			ResNet101-AAVER \cite{aaver} &88.97 &61.18  & ICCV 2019\\
			MRL \cite{mrl} & 87.70 &71.40   & ICME 2019\\
			Fusion-Net \cite{fusionnet} &87.31 &62.40   &IEEE TIP 2019\\
			Mob.VFL-LSTM \cite{vrl} &87.18 &58.08   & ICIP 2019\\
			MGL \cite{mgl}                            &86.10 &65.00    & ICIP 2019\\
			EALN \cite{ealn} &84.39 &57.44   &IEEE TIP 2019\\
			FDA-Net \cite{wild} & 49.43 &N/A  & CVPR 2019\\
			JDRN+Re-ranking  \cite{jdrn} & N/A &73.10    & CVPRW 2019\\
			\hline
			MAD+STR \cite{mattribute}                                            & 89.27&61.11  & ICIP 2018\\
			RAM \cite{ram}                                                                   &88.60&61.50  & ICME 2018\\
			VAMI \cite{vami}                                                             &85.92&61.32  & CVPR 2018\\
			GSTE \cite{gste}                                                                &N/A &59.47     & IEEE TMM 2018\\
			SDC-CNN \cite{sdccnn}                                                  &83.49&53.45  & ICPR 2018\\
			PROVID \cite{veri2017}                                                     &81.56&53.42  & IEEE TMM 2018\\
			NuFACT+Plate-SNN \cite{veri2017}                          &81.11&50.87   & IEEE TMM 2018\\
			SCCN-Ft+CLBL-8-Ft \cite{dhmi}                                          &60.83&25.12  & IEEE TIP 2018\\
			ABLN-Ft-16 \cite{abln}                                                         &60.49&24.92  & WACV 2018\\
			NuFACT \cite{veri2017}                                                     &76.76&48.47  & IEEE TMM 2018\\
			\hline
			VST Path Proposals \cite{vst}                                      &83.49&58.27  & ICCV 2017\\ %
			OIFE+ST \cite{oife}                                                         &68.30&51.42  & ICCV 2017\\
			DenseNet121 \cite{dense}                                          &80.27&45.06   & CVPR 2017\\
			\hline
			FACT \cite{veri2016}                                                    &52.21&18.75   & ICME 2016\\
			VGG-CNN-M-1024 \cite{drdl}                                      &44.10&12.76  & CVPR 2016\\
			GoogLeNet \cite{largecar}                                             &52.32&17.89  & CVPR 2016\\
			\hline
		\end{tabular}
	\end{center}
	\vspace{-0.5cm}
\end{table}

\begin{table}[htp]
	\caption{The performance (\%) comparison on VehicleID.
		 The {\color{red}{red}}, {\color{green}{green}} and {\color{blue}{blue}} rows respectively represent the {\color{red}{$1$st}}, {\color{green}{$2$nd}} and {\color{blue}{$3$rd}} places, according to R1 comparison.}\label{tab:vehicleid}
	\vspace{-.4cm}
	\begin{center}
		\footnotesize
		\setlength{\tabcolsep}{0.5pt}
		\begin{tabular}{lcc cc ccr}
			\hline
			\multirow{2}{*}{Methods}
			&\multicolumn{2}{c} {Test800}
			& \multicolumn{2}{c} {Test1600}
			& \multicolumn{2}{c}{Test2400}
			&\multirow{2}{*}{References}\\
			& R1     &mAP
			& R1      &mAP
			& R1      &mAP \\
			\hline
			\color{red}{HPGN}
			&\color{red}{83.91}&\color{red}{89.60}
			&\color{red}{79.97}&\color{red}{86.16}
			&\color{red}{77.32}&\color{red}{83.60}
			&\color{red}{Proposed}\\
			
			%
			
			QD-DLF \cite{qddlf}
			&{72.32} &{76.54}
			&{70.66}&{74.63}
			&{64.14} &{68.41}
			&IEEE ITS 2020\\
			\hline
			\color{green}{Appearance+License \cite{app}}
			&\color{green}{79.5} &\color{green}{82.7}
			&\color{green}{76.9} &\color{green}{79.9}
			&\color{green}{74.8} &\color{green}{77.7}
			&\color{green}{ICIP 2019}\\
			\color{blue}{MGL \cite{mgl}}
			&\color{blue}{79.6} &\color{blue}{82.1}
			&\color{blue}{76.2} &\color{blue}{79.6}
			&\color{blue}{73.0} &\color{blue}{75.5}
			&\color{blue}{ICIP 2019}\\
			Part Regularization \cite{partreg}
			&{78.40} &{N/A}
			&{75.00}&{N/A}
			&{74.20}&{N/A}
			&CVPR 2019\\
			PRN\cite{prn}
			&78.92 &N/A
			&74.94 &N/A
			&71.58 & N/A
			&CVPRW 2019\\
			Triplet Embedding \cite{triemd}
			&78.80  &86.19
			&73.41  &81.69
			&69.33  &78.16
			& IJCNN 2019\\
			
			MRM \cite{mrm}
			&76.64 &80.02
			&74.20 &77.32
			&70.86 &74.02
			&Neurocomputing 2019\\
			XG-6-sub-multi \cite{xg6}
			&76.1 &N/A
			&73.1 &N/A
			&71.2 &N/A
			& IEEE ITS 2019\\
			GRF+GGL \cite{ggl}
			&77.1&N/A
			&72.7&N/A
			&70.0&N/A
			&IEEE TIP 2019\\

			MSV \cite{msv}
			&75.1 &79.3
			&71.8 &75.4
			&68.7 &73.3
			&ICASSP 2019\\
			
			DQAL \cite{tvt}
			&74.74   &N/A
			&71.01  &N/A
			&68.23  &N/A
			&IEEE TVT 2019\\
			
			EALN \cite{ealn}
			&75.11 &77.5
			&71.78 &74.2
			&69.30 &71.0
			& IEEE TIP 2019\\
			Mob.VFL-LSTM \cite{vrl}
			&73.37 &N/A
			&69.52 &N/A
			&67.41 &N/A
			& ICIP 2019\\
			
			%
			
			ResNet101-AAVER \cite{aaver}
			&74.69 &N/A
			&68.62 &N/A
			&63.54 &N/A
			& ICCV 2019\\

			TAMR \cite{tamr}
			&66.02 &N/A
			&62.90 &N/A
			&59.69 &N/A
			&IEEE TIP 2019\\
			MLSR \cite{mlsr}
			&65.78 &N/A
			&64.24 &N/A
			&60.05 &N/A
			& Neurocomputing 2019\\
			RPM \cite{rpm}
			&65.04&N/A
			&62.55&N/A
			&60.21&N/A
			&ICMEW 2019\\
			{SFF+SAtt \cite{sff}}
			&64.50 &N/A
			&59.12 &N/A
			&54.41 &N/A
			&IJCNN 2019\\
			FDA-Net \cite{wild}
			&N/A &N/A
			&59.84 & 65.33
			&55.53 & 61.84
			&CVPR 2019\\

			\hline
			GSTE \cite{gste}
			&75.90     &75.40
			&74.80     &74.30
			&74.00     &72.40
			&IEEE TMM 2018\\
			RAM \cite{ram}
			&{75.20} &{N/A}
			&{72.3}&{N/A}
			&{67.70}&{N/A}
			&ICME 2018\\
			C2F \cite{c2f}
			&61.10        &63.50
			&56.20     &60.00
			&51.40  &53.00
			&AAAI 2018\\
			VAMI \cite{vami}
			&63.12  &N/A
			&52.87      &N/A
			&47.34 &N/A
			&CVPR 2018\\
			SDC-CNN \cite{sdccnn}
			&{56.98} &63.52
			&{50.57}   &57.07
			&{42.92} &49.68
			&ICPR 2018\\

			NuFACT \cite{veri2017}
			&48.90              &N/A
			&43.64             &N/A
			&38.63             &N/A
			&IEEE TMM 2018\\
			
			MAD+STR \cite{mattribute}
			&N/A     &82.20
			&N/A      &75.90
			&N/A         &72.80
			&ICIP 2018\\
			
			PMSM \cite{pmsm}
			&N/A &64.20
			&N/A &57.20
			&N/A &51.80
			&ICPR 2018\\
			
			MSVF \cite{vric}
			&N/A      &N/A
			& N/A& N/A
			& 46.61  &N/A
			&GCPR 2018\\

			ABLN-32 \cite{abln}
			&52.63            &N/A
			&N/A& N/A
			&N/A& N/A
			&WACV 2018\\

			\hline
			DJDL \cite{djdl}
			&72.30        &{N/A}
			&70.80         &{N/A}
			&68.00       &{N/A}
			&ICIP 2017\\
			Improved Triplet \cite{imtri}
			&69.90    &{N/A}
			&66.20    &{N/A}
			&63.20    &{N/A}
			&ICME 2017\\
			DenseNet121 \cite{dense}
			&{66.10}  &68.85
			&{67.39}      &{69.45}
			&{63.07}   &{65.37}
			&CVPR 2017\\
			MGR\cite{mgr}
			&N/A     &62.80
			&N/A    &62.30
			&N/A      &61.23
			&ICCV 2017\\
			OIFE+ST \cite{oife}
			&N/A       &N/A
			&N/A       &N/A
			&67.00 &N/A
			&ICCV 2017\\

			\hline
			DRDL \cite{drdl}
			&48.91                 &N/A
			&46.36          &N/A
			&40.97         &N/A
			&CVPR 2016\\
			FACT \cite{veri2016}
			&49.53               &N/A
			&44.63               &N/A
			&39.91             &N/A
			&ICME  2016\\
			\hline
		\end{tabular}
	\end{center}
\end{table}

\subsubsection{Comparisons on VeRi776}
    From Table \ref{tab:veri776}, it can be found that the vehicle re-identification performance has made significant progress between 2016 and 2020 on VeRi776 \cite{veri776}. Especially in 2019, many approaches \cite{app,sff,partreg,pamtri,mlfn,mtml,mrm,dmml,triemd,cityflow} acquire more than 90\% rank-1 identification rates. Under this background, the proposed HPGN method achieves the highest rank-1 identification rate (i.e., 96.72\%) and the largest mAP (i.e., 80.18\%). For example, among those compared state-of-the-art methods, rank-1 identification rates of the 2nd place (i.e., Appearance+License \cite{app}) and the 3rd place (i.e., SFF+SAtt+TBR \cite{sff}) are 1.31\% and 1.79\% lower than those of HPGN, respectively. Besides, regarding mAP performance, Appearance+License \cite{app} and SFF+SAtt+TBR \cite{sff} are 2.10\% and 6.07\% lower than that of HPGN, respectively. It is worth mentioning that both Appearance+License \cite{app} and SFF+SAtt+TBR \cite{sff} use the same backbone network (i.e., ResNet-50 \cite{resnet}) to HPGN. However, HPGN does not require vehicle attribute aids, e.g., license plates and temporal Bayesian re-rankings (TBR).

\subsubsection{Comparisons on VehicleID}
    Table \ref{tab:vehicleid} shows the comparison of the proposed HPGN method and state-of-the-art approaches on VehicleID \cite{drdl} that has a larger data scale than VeRi776 \cite{veri776}.  One can see that many approaches (e.g., \cite{app,mgl,partreg,triemd,prn,mrm,xg6,ggl,msv,tvt,ealn}) made significant progress in 2019 on VehicleID, but HPGN outperforms all of them. More specifically, Appearance+License \cite{app} still is the 2nd place and in superior to HPGN. For example, the R1 and mAP of Appearance+License \cite{app} are 4.41\% and 6.90\% lower than those of HPGN on Test800, respectively. Regarding the 3rd place, without the help of temporal Bayesian re-ranking (TBR), the performance rank of SFF+SAtt \cite{sff} drops dramatically. MGL \cite{mgl} replaces SFF+SAtt \cite{sff} as the 3rd place, and its R1 and mAP are 4.31\% and 7.50\% lower than those of HPGN on Test800, respectively. In summary, although Appearance+License \cite{app}, SFF+SAtt \cite{sff}, MGL \cite{mgl}, and HPGN adopt the same backbone network (i.e., ResNet-50 \cite{resnet}), HPGN acquires the 1st place on both VehicleID \cite{drdl} and VeRi776 \cite{veri776}, which solidly demonstrates the HPGN's superiority.

\subsubsection{Comparisons on VeRi-Wild}
   Table \ref{tab:wild} shows the comparison result on VeRi-Wild \cite{wild}. Since VeRi-Wild \cite{wild} is a newly released dataset, only a few methods have been evaluated on it. Among those methods, it can be found that the proposed HPGN method obtains the 1st place again. For example, HPGN outperforms the 2nd place, namely, Triplet Embedding \cite{triemd}, by a 12.69\% higher rank-1 identification rate on Test10000.

\begin{table}[tp]
	\caption{The performance (\%) comparison on VERI-Wild. The {\color{red}{red}}, {\color{green}{green}} and {\color{blue}{blue}} rows respectively represent the {\color{red}{$1$st}}, {\color{green}{$2$nd}} and {\color{blue}{$3$rd}} places, according to the R1 comparison.}\label{tab:wild}	
	\begin{center}
		\footnotesize
		\setlength{\tabcolsep}{1pt}
		\begin{tabular}{lcc cc ccr}
			\hline
			\multirow{2}{*}{Methods}
			&\multicolumn{2}{c} {Test3000}
			&\multicolumn{2}{c} {Test5000}
			&\multicolumn{2}{c}{Test10000}
			&\multirow{2}{*}{References}\\
			& R1     &mAP
			& R1      &mAP
			& R1      &mAP \\
			\hline
			\color{red}{HPGN}
			&\color{red}{91.37}&\color{red}{80.42}
			&\color{red}{88.21}&\color{red}{75.17}
			&\color{red}{82.68}&\color{red}{65.04}
			&\color{red}{Proposed}\\
			
			
			\color{green}{Triplet Embedding \cite{triemd}}
			&\color{green}{84.17} &\color{green}{70.54}
			&\color{green}{78.22} &\color{green}{62.83}
			&\color{green}{69.99} &\color{green}{51.63}
			&\color{green}{IJCNN 2019}\\
			
			\color{blue}{FDA-Net} \cite{wild}
			&\color{blue}{64.03} &\color{blue}{35.11}
			&\color{blue}{57.82} &\color{blue}{29.80}
			&\color{blue}{49.43} &\color{blue}{22.80}
			&\color{blue}CVPR 2019\\
			{GSTE} \cite{gste}
			&{60.46} &{31.42}
			&{52.12} &{26.18}
			&{45.36} &{19.50}
			&IEEE TMM 2018\\
			Unlabled GAN \cite{unlabledgan}
			&58.06 &29.86
			&51.58 &24.71
			&43.63 &18.23
			&ICCV 2017\\
			GoogLeNet \cite{googlenet}
			&57.16 &24.27
			&53.16 &24.15
			&44.61 &21.53
			&CVPR 2015\\
			HDC \cite{hdc}
			&57.10 &29.14
			&49.64 &24.76
			&43.97 &18.30
			&ICCV 2017\\
			DRDL \cite{drdl}
			&56.96 &22.50
			&51.92 &19.28
			&44.60 &14.81
			&CVPR 2016\\
			Softmax \cite{veri776}
			&53.40 &26.41
			&46.16 &22.66
			&37.94 &17.62
			&ECCV 2016\\
			Triplet \cite{facenet}
			&44.67 &15.69
			&40.34 &13.34
			&33.46 &9.93
			&CVPR 2015\\
			\hline
		\end{tabular}
	\end{center}
\end{table}

\subsection{Analysis}
   The comparison results presented in Tables \ref{tab:veri776}, \ref{tab:vehicleid}, and \ref{tab:wild} demonstrate that the proposed HPGN method is superior to a lot of state-of-the-art vehicle re-identification methods. In what follows, the proposed HPGN method is comprehensively analyzed from four aspects to investigate the logic behind its superiority. (1) The role of pyramidal graph network (PGN). (2) The influence of changing depths of spatial graph networks (SGNs). (3) The impact of varying HPGN's scale configurations. (4) The spatial significance visualization.

\begin{table*}[htp]
	\caption{ The role (\%) of pyramidal graph network (PGN).}
	\label{tab:role_pgn}
	\begin{center}
		\setlength{\tabcolsep}{7pt}
		\begin{tabular}{c|cc | cc cc cc | cc cc cc}
			\hline
			\multirow{3}{*}{Methods}
			& \multicolumn{2}{c|}{\multirow{2}{*}{VeRi776}}&\multicolumn{6}{c|}{VehicleID} &\multicolumn{6}{c}{VERI-Wild}\\
			& &
			& \multicolumn{2}{c}{Test800}
			& \multicolumn{2}{c}{Test1600}
			& \multicolumn{2}{c|}{Test2400}
			& \multicolumn{2}{c}{Test3000}
			& \multicolumn{2}{c}{Test5000}
			& \multicolumn{2}{c}{Test10000}\\
			&R1      &mAP
			&R1      &mAP
			&R1      &mAP
			&R1      &mAP
			&R1      &mAP
			&R1      &mAP
			&R1      &mAP\\
			
			\hline
			
			HPGN 
			&\textbf{96.72} &\textbf{80.18}
			&\textbf{83.91} &\textbf{89.60}
			&\textbf{79.97} &\textbf{86.16}
			&\textbf{77.32} &\textbf{83.60}
			&\textbf{91.37} &\textbf{80.42}
			&\textbf{88.21} &\textbf{75.17}
			&\textbf{82.68} &\textbf{65.04} \\
			
			%
			\color{black}{HPGN$_{\rm{OI}}$}
			&\color{black}{96.06}&\color{black}{78.87}
			&\color{black}{82.45}&\color{black}{88.85}
			&\color{black}{78.62}&\color{black}{85.28}
			&\color{black}{75.88} &\color{black}{82.55}
			&\color{black}{90.11}&\color{black}{78.22}
			&\color{black}{86.03}&\color{black}{74.31}
			&\color{black}{79.91} &\color{black}{63.27}      \\

			HPGN$_{\rm{NG}}$
			&95.23 &75.90
			&82.02 &87.87
			&78.04 &84.08
			&75.23 &81.65
			&88.67 &76.34
			&85.84 &72.48
			&78.32 &61.22        \\

			Baseline
			&94.12 &74.65
			&79.48 &86.13
			&75.81 &82.21
			&72.80 &79.54
			&88.18 &72.16
			&83.95 &65.08
			&76.82 &55.60             \\
			\hline
		\end{tabular}
	\end{center}
\end{table*}

\begin{table*}[htp]
	\caption{The performance  (\%) comparison among those hybrid pyramidal graph networks (HPGNs) holding pyramidal graph networks (PGNs) of different scale configurations.}
	\label{tab:scale}
	\begin{center}
		\setlength{\tabcolsep}{5pt}
		\begin{tabular}{cc|cc |cc cc cc |cc cc cc}
			\hline
			\multirow{3}{*}{Names}
			&
			\multirow{3}{*}{Scales}
			& \multicolumn{2}{c|}{\multirow{2}{*}{VeRi776}}
			&\multicolumn{6}{c}{VehicleID}&\multicolumn{6}{|c}{VERI-Wild}\\
			& & &
			& \multicolumn{2}{c}{Test800}
			& \multicolumn{2}{c}{Test1600}
			& \multicolumn{2}{c}{Test2400}
			& \multicolumn{2}{|c}{Test3000}
			& \multicolumn{2}{c}{Test5000}
			& \multicolumn{2}{c}{Test10000}\\
			&
			& R1      &mAP
			& R1      &mAP
			& R1      &mAP
			& R1      &mAP
			& R1      &mAP
			& R1      &mAP
			& R1      &mAP\\
			\hline
			Baseline &S5 
			&94.12 &74.65
			&79.48 &86.13
			&75.81 &82.21
			&72.80 &79.54
			&88.18 &72.16
			&83.95 &65.08
			&76.82 &55.60  \\
			HPGN1 &S5, S4
			&95.65 &78.91
			&81.82 &87.93
			&77.49 &83.96
			&74.56 &81.12
			&89.43 &76.52
			&86.07 &72.42
			&79.37 &61.25 \\
			HPGN2 &S5, S4, S3
			&95.95 &79.03
			&82.92 &88.75
			&79.18 &85.40
			&75.96 &82.47
			&89.71 &77.34
			&86.82 &73.88
			&79.96 &62.45 \\
			HPGN3 &S5, S4, S3, S2
			&96.06 &79.35
			&83.58 &89.16
			&79.33 &85.54
			&76.05 &82.57
			&90.66 &78.84
			&87.13 &74.09
			&80.24 &63.88 \\
			HPGN &S5, S4, S3, S2, S1
			&\textbf{96.72} &\textbf{80.18}
			&\textbf{83.91} &\textbf{89.60}
			&\textbf{79.97} &\textbf{86.16}
			&\textbf{77.32} &\textbf{83.60}
			&\textbf{91.37} &\textbf{80.42}
			&\textbf{88.21} &\textbf{75.17}
			&\textbf{82.68} &\textbf{65.04} \\
			\hline
		\end{tabular}
	\end{center}
\end{table*}

\begin{figure}[tp]
	\centering
	\includegraphics[width=.95\linewidth]{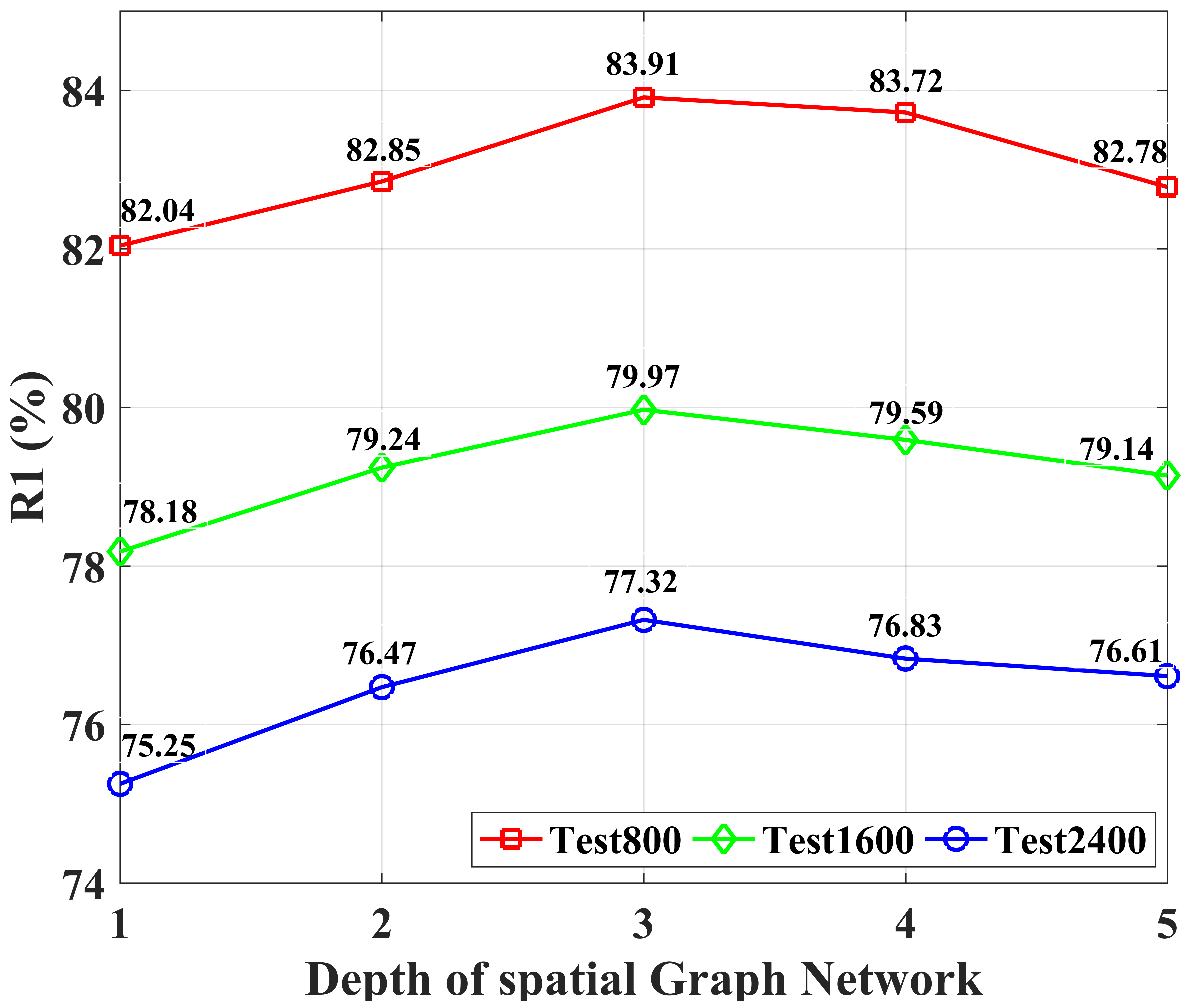}
	\caption{The rank-1 identification rate (R1) comparison among spatial graph networks of different depth on VehicleID.}
	\label{fig:r1}
\end{figure}

\begin{figure}[tp]
	\centering
	\includegraphics[width=.95\linewidth]{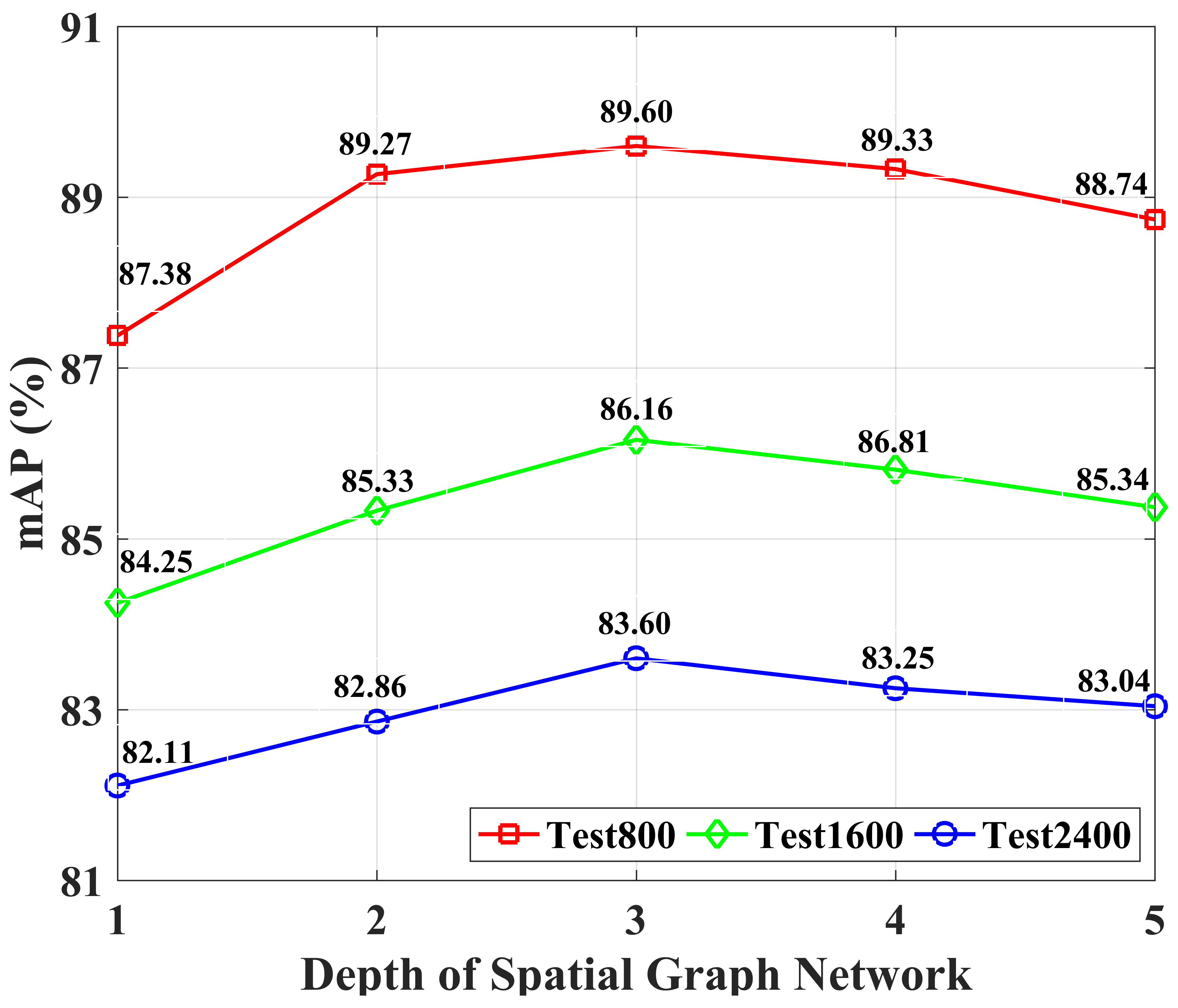}
	\caption{The mean average precision (mAP) comparison among spatial graph networks of different depths on VehicleID.}
	\label{fig:map}
\end{figure}

\subsubsection{Role of PGN}
	 To validate the role of PGN, HPGN is compared with a baseline method that replaces the PGN with a global max pooling (GMP) layer while holds the same backbone network to HPGN. As shown in Table \ref{tab:role_pgn}, HPGN significantly outperforms the baseline method. For example, compared to the baseline method, on VeRi776 \cite{veri776}, HPGN holds a larger 5.53\% mAP, and on Test2400 of VehicleID \cite{drdl}, HPGN holds a larger 4.06\% mAP. Those comparisons overall demonstrate the PGN can significantly improve vehicle re-identification.
	
	Considering that PGN includes two essential designs, i.e., the spatial graph network (SGN) and the pyramidal architecture (PA), more ablations are investigated. Two different simplified HPGN, i.e., HPGN$_{\rm{OI}}$ and HPGN$_{\rm{NG}}$, are constructed. HPGN$_{\rm{OI}}$ and HPGN$_{\rm{NG}}$ also use the same backbone network to HPGN. Besides, they hold a pyramidal architecture (PA) containing the same scale configuration to that of HPGN. However, the differences are as follows. (1) HPGN$_{\rm{OI}}$ \textbf{O}nly retains a node \textbf{I}tself and gives up aggregating the node's spatial neighbors (i.e., setting $S = \bm{0}$ in Eq. \eqref{eq:agg}) during the spatial graph network's propagation. (2) HPGN$_{\rm{NG}}$ contains \textbf{N}o spatial \textbf{G}raph network, that is, it entirely disregards spatial graph networks in its pyramidal architecture.
	
	Comparing HPGN$_{\rm{OI}}$, HPGN$_{\rm{NG}}$, and HPGN, one can observe that HPGN$_{\rm{OI}}$ is better than HPGN$_{\rm{NG}}$ from Table \ref{tab:role_pgn}, demonstrating the beneficial effect of re-weighting nodes. Moreover, HPGN outperforms HPGN$_{\rm{OI}}$, which suggests that aggregating both a node and the node's neighbors can feed more information to improve the effect of re-weighting nodes. Besides, HPGN$_{\rm{NG}}$ is superior to the Baseline that replaces the PGN with a global max pooling (GMP) layer, which shows that the plain pyramidal architecture containing no SGN still is helpful to improve vehicle re-identification.

\subsubsection{Influence of Changing SGN's Depths}
     Fig. \ref{fig:r1} shows the rank-1 identification rate (R1) comparison among SGNs of different depths (i.e., numbers of spatial graphs), and Fig. \ref{fig:map} presents the mAP comparison among those SGNs. It can be observed that both R1 and mAP performance fluctuate with SGN's depths, and the best R1 and mAP are acquired when each SGN's depth is set to 3. Furthermore, one can see even the worst results (i.e., those R1s and mAPs resulted from setting each SGN's depth to 1) are comparable with the results of HPGN$_{\rm{NG}}$ (see Table \ref{tab:role_pgn}). These comparison results demonstrate the beneficial role of using SGNs again.

\subsubsection{Impact of Varying HPGN's Scale Configurations}
    In addition to the Baseline and HPGN showed in Table \ref{tab:role_pgn}, three HPGNs of different scale configured PGNs, namely, HPGN1, HPGN2, and HPGN3, are further evaluated. As shown in Table \ref{tab:scale}, HPGN1 represents that the applied PGN contains S5 and S4 scales. HPGN2 denotes that the used PGN includes S5, S4, and S3 scales. HPGN3 represents that the adopted PGN contains S5, S4, S3, and S2 scales. The HPGN applies a PGN that has S5, S4, S3, S2, and S1 scales. The Baseline only uses the S5 scale. The structures of S1, S2, S3, S4, and S5 can be found in Fig. \ref{fig:framework}. All these methods use the same backbone network.

    From Table \ref{tab:scale}, it can be observed that the more scales applied in a PGN, the better performance will be acquired on both VeRi776 \cite{veri776}, VehicleID \cite{drdl}, and VeRi-Wild \cite{wild}. For example, on VeRi776 \cite{veri776}, HPGN's mAP is 5.53\%, 1.27\%, 1.15\%, and 0.83\% higher than that of Baseline, HPGN1, HPGN2, and HPGN3, respectively. On the largest testing subset (i.e., Test2400) of VehicleID \cite{drdl}, HPGN defeats Baseline, HPGN1, HPGN2, and HPGN3 by 4.52\%, 2.76\%, 1.36\%, and 1.27\% higher rank-1 identification rates, respectively. These results demonstrate that the proposed HPGN exploring spatial significance of feature maps at multiple scales helps promote vehicle re-identification performance.

\begin{figure}[tp]
	\centering
	\includegraphics[width=1\linewidth]{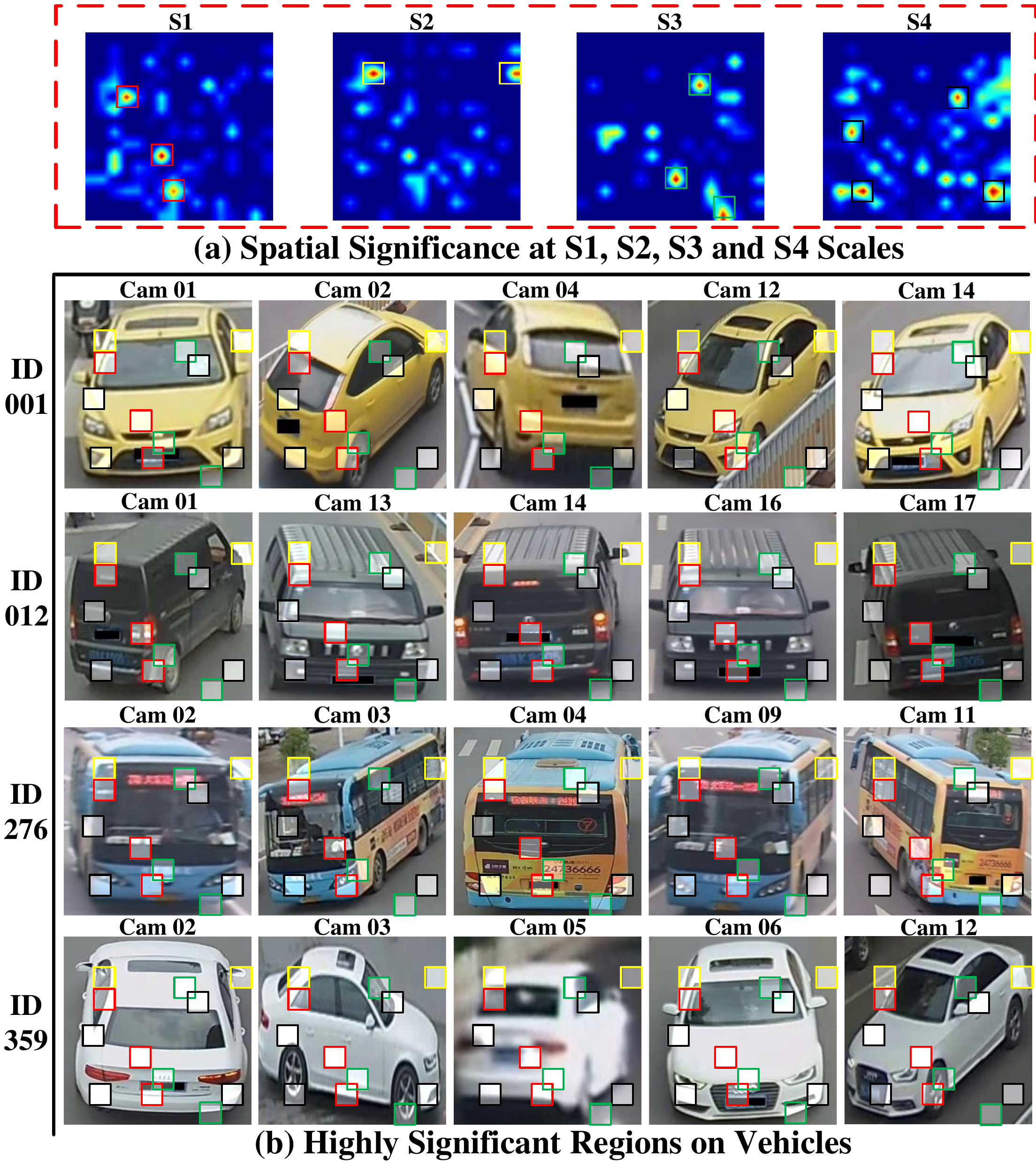}
	\caption{Visualization results on VeRi776.
	}
	\label{fig:vis}
\end{figure}

\subsubsection{Spatial Significance Visualization}
	The spatial significance visualization is realized by using the gradient weighted class activation mapping (Grad-CAM) \cite{gradcam} method to visualize the PGN's learnable parameter matrixes (i.e., $\varTheta$ in Eq. \eqref{eq:dot}), as shown in Fig. \ref{fig:vis} (a), where the higher significant regions are rendered in hotter colors (e.g., red), while the lower significant regions are rendered in cool colors (e.g., blue). The S1, S2, S3 and S4 scale configurations can be found in Fig. \ref{fig:framework}.

	Two interesting observations can be found. First, at each scale, highly significant regions appear scatteredly, which demonstrates that the global pooling operation equally treats each region will lose the useful spatial significance information. Second, at four different scales, highly significant regions appear at different locations, which shows that there is a complementation of exploring spatial significance at multiple scales.
	
	To further illustrate spatial significance, those highly significant regions explored in Fig. \ref{fig:vis} (a) are served as observation windows on vehicle images captured from different cameras. As shown in Fig. \ref{fig:vis} (b), informative parts (e.g., lightings, license plates, wheels, stickers or pendants on windshields) of vehicles are frequently observed, demonstrating the proposed HPGN method can explore spatial significance effectively.

\section{Conclusion}\label{sec:con}
		In this paper, a hybrid pyramidal graph network (HPGN) is proposed for vehicle re-identification. The HPGN consists of a backbone network (i.e., ResNet-50) and a novel pyramidal graph network (PGN). The backbone network is applied to extract feature maps from vehicle images, the PGN is designed to work behind the backbone network to comprehensively explore spatial significance of the feature maps at multiple scales. The key module of PGN is the innovative spatial graph network (SGN), which stacks a list of spatial graphs (SGs) on feature maps of a specific scale and explore the spatial significance via re-weighting nodes during the propagation. Experiments on three large scale databases demonstrate that the proposed HPGN method outperforms state-of-the-art vehicle re-identification approaches.

\ifCLASSOPTIONcaptionsoff
  \newpage
\fi

\bibliographystyle{IEEEtran}
\small{
\bibliography{reffullv2}
}

\end{document}